# A Machine Learning Method for Material Property Prediction: Example Polymer Compatibility

Zhilong Liang,[1] Zhiwei Li,[2] Shuo Zhou,[2,3] Yiwen Sun,[1,3] Jinying Yuan,[2,*] and Changshui Zhang [1,4,**]

[1] Institute for Artificial Intelligence, Tsinghua University (THUAI),
  State Key Lab of Intelligent Technologies and Systems,
  Beijing National Research Center for Information Science and Technology (BNRist),
  Department of Automation, Tsinghua University, Beijing, P. R. China

[2] Key Lab of Organic Optoelectronics and Molecular Engineering of Ministry of Education,
  Department of Chemistry, Tsinghua University, Beijing, P. R. China.

[3] Both authors share the same contribution

[4] Lead Contact

*Correspondence: yuanjy@tsinghua.edu.cn

**Correspondence: zcs@mail.tsinghua.edu.cn

February 27，2022

**SUMMARY**

Prediction of material property is a key problem because of its significance to material design and screening. We present a brand-new and general machine learning method for material property prediction. As a representative example, polymer compatibility is chosen to demonstrate the effectiveness of our method. Specifically, we mine data from related literature to build a specific database and give a prediction based on the basic molecular structures of blending polymers and, as auxiliary, the blending composition. Our model obtains at least 75% accuracy on the dataset consisting of thousands of entries. We demonstrate that the relationship between structure and properties can be learned and simulated by machine learning method.

**INTRODUCTION**

Prediction of material property is highly significant for discovering new materials. Traditional empirical trial and error method is generally high time and labor consuming, considering the nearly infinite material structures and the required expertise [1]. Density functional theory (DFT)-based method and molecular dynamics simulation provides a new paradigm for studying the properties of materials. DFT is successfully applied to prediction of complex system behavior at an atomic scale and molecular dynamics simulation helps to predict material structure and properties of thermodynamics [2,3]. Nevertheless, numerical algorithms and computing resources become the main bottleneck when it comes to macroscale and multibody system.

Machine Learning (ML) which is data-driven achieves great progress on various tasks, such as image recognition [4,5] and natural language processing [6,7]. Recently, scientists start to apply machine learning to material research and property prediction. For biological macro molecule material, ML method achieves high accuracy in protein structure AlphaFold2 network [8]. In inorganic material field, ML assists prediction of low-dimensional materials' properties [9], ceramics properties [10], steel fatigue strength [11], alloy miscibility [12], battery electrolytes materials [13] and photovoltaic materials [14], etc. Researchers also utilize machine learning method to research properties of metal-organic frameworks (MOF) and establish some specific databases [15,16,17]

Except as mentioned above, machine learning could especially help to research polymer materials. Polymer materials have been widely used in electrical, medicine and all aspects of production, engineering and daily life because of their excellent properties. The properties of polymer materials are closely related to the multidimensional factors in the process of polymer synthesis and processing, which makes it complicated and challengeable to give accurate prediction [18,19,20,21]. Researchers have made active and effective attempts to apply ML method in exploring polymer syntheses and polymer materials [22]. The copolymer synthesis and defectivity [23,24,25], mechanical properties of polymer composites [26], liquid crystal behavior of copolyether [27], thermal conductivity [28], dielectric properties [29], glass transition, melting, and degradation temperature and quantum physical and chemical properties [30,31,32,33] have been applied with machine learning and good prediction accuracy is achieved. Muramatsu et al. [34] have used ML method to investigate the relationship between the phase separation structure of polymer blend and Young's modulus, and builds a predictive framework based on two-dimensional images of polymer blend as the descriptor.

Herein, we apply a brand-new and general ML method to material property prediction. As a start for prediction of a series of work on polymer structure and property, we choose polymer blend compatibility as our focus, since polymer blend materials are able to integrate the advantages of various polymers and play an important role in industry. Trunk of our method includes: (1) Building up a framework to mine data from literature and construct a specific dataset. (2) Designing a specialized prediction model, Half Dense Difference Network. It is noteworthy that we can easily generalize the whole procedure to any other material field. Following, we will give a brief introduction to polymer compatibility and prediction methods.

Polymer compatibility is a key physical quantity to influence properties of polymer blend. Polymer compatibility means the total miscibility of homo-polymers and of random copolymers with each other on a molecular scale [35]. Poor compatibility will severely limit the utility of polymer blend so scientists generally prefer polymer combinations with good compatibility and search for them through series of chemical theories.

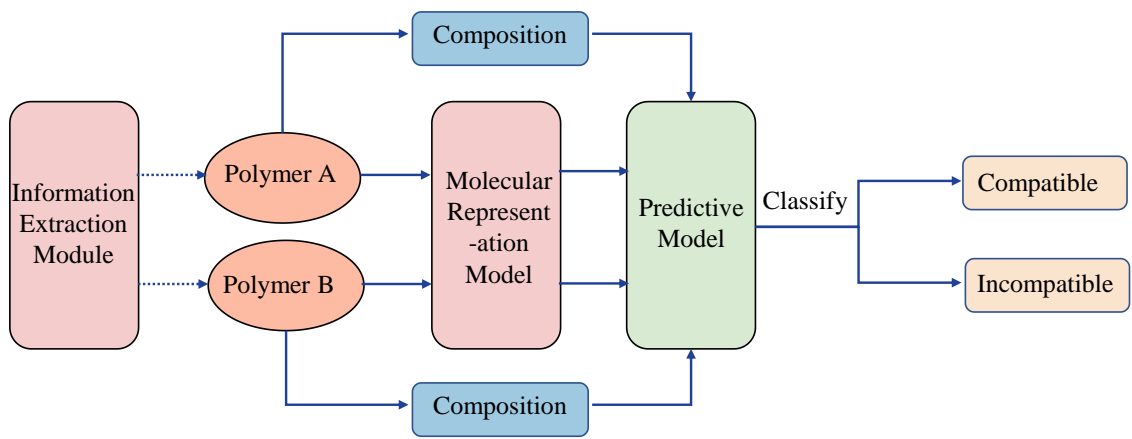

**Figure 1. Whole Information Flow of Prediction Method**
For binary polymer blend, we collect these fundamental information, repeating units and component. We transform such information to vector representations, and then pass them to Predictive Model to classify them as compatible or incompatible.

Some theories are conducted to predict polymer blend compatibility. One of the most fundamental model of polymer blends is developed by Flory and Huggins, which theoretically describes the thermodynamic effect when polymers are mixed [36]:

$$\frac{\Delta G_M}{RT} = n_1 \ln\phi_1 + n_2 \ln\phi_2 + n_1 \phi_2 \chi_{12}, \tag{1}$$

where $n_i$ means the number of moles of component $i$, $\phi_i$ means volume fraction of component $i$, $\chi_{12}$ means interaction parameter, $R$ is the gas constant and $T$ is the absolute temperature.

According to the equation, the Flory–Huggins (F-H) interaction parameter, $\chi$ is considered as a characterizing factor to the compatibility of polymer blend [37]. Methods to estimate $\chi$ can be classified into two categories: experimental methods [38,39,40] and numerical simulation computation methods [41,42,43]. Compared with these estimation methods, approximate prediction method based upon Hildebrand Solubility Parameters (HSP) is shown as follow [44]:

$$\chi_{12} = \frac{V}{RT}(\delta_1 - \delta_2)^2, \tag{2}$$

where $V$ means actual volume of a polymer segment, $\delta_i$ means Hildebrand Solubility Parameters of component $i$. Since $\chi_{12}$ indicates the interaction parameter mentioned in F-H theory, the more similar $\delta$ parameters are, the more possibly polymer blends are compatible [45,46]. Askadskii tries to predict polymer compatibility in the use of HSP criterion [47], and makes efforts to predict compatibility directly from chemical structure of the repeating units of polymers [48]. Schneier precisely classifies 28 specific polymer blends (96.43% accuracy) [49] through modified HSP-calculation method and a critical value of 0.010 cal/mol. Although the generalizability is not satisfying enough, the attempt to predict compatibility utilizing repeating unit inspires us greatly.

Besides theoretical research, several researchers focus on specific compatibility experiments of polymer combination. These data are reported in thousands of articles. We utilize this information dispersed in sporadic articles to construct a specific database. Based on it, we design a predictive ML model and receive good results. Whole process is roughly shown as Figure 1. In summary, our main contributions can be listed as follows:
- We design a scheme to mine data from literature and use the scheme to build a specific polymer compatibility database.
- We present our model Half Dense Difference Network (HDDN). We show that it is possible to predict polymer compatibility with only repeating structure and composition by machine learning method.
- We show that it is possible to verify and discover new chemical and material science principles via the help of machine learning model.

### RESULTS

**Data Collection and Information Extraction**

Although many researchers conduct lots of compatibility studies and publish their work, there is not a specific database constructed for polymer compatibility. We collect data by following means.
- Database extraction: Database PoLyInfo is developed by National Institute for Materials Science (NIMS) [50]. It contains a number of polymer blend information and blend morphology information. Some entries have clear compatibility information which can be inferred from morphology description, such as miscible, compatible, incompatible and so on. We collect them and tag them with compatible and incompatible according to morphology description. We discard those cases where blend is partially compatible or description is ambiguous.
- Text Data Mining: We search and download papers related to keywords 'Compatibility' and 'Polymer' from Google Scholar and Tsinghua University Library, which sum up to 47K articles. In these articles, some sentences contain clear compatibility information. For example, "Results of physical properties measurements reveal the blends of SR and FKM are technologically compatible". We design a filter to automatically export these sentences from articles, and get the database from literature.

All entries we collect above are transformed from text to standard vector form consisting of polymers repeating unit structure, component information and compatibility label. Our ultimate dataset contains 1.4K reliable entries. We divide these data into training-set, valid-set and test-set in two ways. (1) Random Division: we randomly choose 64% data for training, 16% for validation and 20% for testing. (2) Balanced Division: we make sure all combinations of polymer blends in test-set don't exist in training-set and valid-set, which guarantees the model learns the rule rather than remember the combinations. At the same time, we balance the categorical proportion of each subset by copying incompatible samples. Among these sets, valid-set is used to choose proper super-parameters and test-set is used to verify generalizability. Details of division are listed in Table1. Data size of each subset are listed and inside parentheses are incompatible rate.

**Table 1. Statistics of Dataset**

| Division | Size | Division Set | | |
|---|---|---|---|---|
| | | Training-set | Valid-set | Test-set |
| Random | 1390 | 889 (38.5%) | 223 (41.7%) | 278 (40.0%) |
| Balanced | 1766 | 1179 (50.0%) | 281 (50.0%) | 306 (50.0%) |

**Competing Results**

Up to now, there is no similar ML model published to solve polymer compatibility prediction. Given that, we apply several fundamental and widely used ML model to this problem. On datasets generated in both ways, we compare our model against several possible competitors. In consideration of fairness, we roughly make numbers of layers and nodes similar.

- Hildebrand Solubility Parameters (HSP): HSP considers segment volume, polymer density and structure features, and reaches over 95% accuracy in Schneier's database. When the $\Delta H_m$ value calculated is not larger than 0.010, the blend is predicted to be compatible, otherwise incompatible. We play this procedure on our dataset, using HSP values from Polymer Database[1] and name them HSP and in the following.
- Multi-Layer Perception (MLP): as mentioned above, MLP is a fundamental neural network model. Through changing the number of layers and nodes and nonlinear function, we can make the MLP fit a function and address some predictive problems. This method replaces our Features Dense Module and Decision Module with MLP.
- Concatenated-Difference-Net (CDN): this model uses another layer-connection method comparing to our Difference Module. CDN concatenates different layers value together and makes the dimension higher, while our model adds features together and guarantees dimension invariant.
- Dense-Net (DN): it is a very popular model in computer vision tasks. Compared with our model, DN contains dense-connection in Difference and Decision Module. We try to compare our model with it and prove that our design according to polymer structure and compatibility mechanism is useful.

**Table 2. Results on Test-set with Competing Models**

| Model | Random Division Results | | | | | |
|---|---|---|---|---|---|---|
| | MSE | Accuracy (%) | Precision (%) | Recall (%) | Specificity (%) | $F_1$ score (%) |
| HSP [a] | —— | 46.91 | 32.88 | 39.34 | 51.49 | 35.82 |
| MLP | **5.09** | 89.98 | **91.16** | 83.00 | **94.61** | 86.85 |
| CDN | 5.55 | 90.65 | 89.56 | 86.71 | 93.26 | 88.10 |
| DN | 5.60 | 89.12 | 86.50 | 86.36 | 90.94 | 86.36 |
| HDDN | 7.25 | **90.89** | 89.75 | **87.24** | 93.31 | **88.43** |
| Model | Balanced Division Results | | | | | |
| | MSE | Accuracy (%) | Precision (%) | Recall (%) | Specificity (%) | $F_1$ score (%) |
| HSP | —— | 68.93 | 70.59 | 66.67 | 71.26 | 65.57 |
| MLP | 7.21 | 63.56 | 67.17 | 51.67 | 75.00 | 57.36 |
| CDN | 6.48 | 65.97 | 65.02 | 67.67 | 64.34 | 65.87 |
| DN | 10.74 | 62.42 | 65.36 | 45.42 | 75.77 | 51.08 |
| HDDN | **6.14** | **75.75** | **76.99** | **74.13** | **77.31** | **74.80** |

[a] HSP method gives prediction based on fixed criterion so has no training process and MSE result.

Results on Random division and Balanced division testset are presented in Table 2. Since incompatible rate is 40% in the random division testset, the model is effective when classification accurate is much higher than 60%. Our HDDN model reaches 90.89% accuracy and highest Recall (87.24%) and $F_1$ score (88.43%). As for other models, MLP, CDN and DN all get high accuracy which is close to HDDN while HSP method obtains accuracy ranging from 31.58% to 46.91%, much lower than Schneier's 96.43%. This is because HSP method is weak in dealing with polar systems, where dipole-dipole interaction or hydrogen bonding further increase compatibility, leading to poor precision. For example, poly(p-hydroxystyrene) is compatible with poly(ethylene oxide), poly(methyl methacrylate), poly(lactic acid), poly(methyl vinyl ether) and poly(vinyl acetate) due to the strong hydrogen bonding, which are falsely predicted. Among all machine learning models, specificity is all obviously higher than recall, which indicates that models perform better when the objects are compatible. This phenomenon can be explained as the random divided dataset is not balanced and compatible samples conquer more proportion (60%), so models pay more attention on this part of data.

---

[1] www.polymerdatabase.com

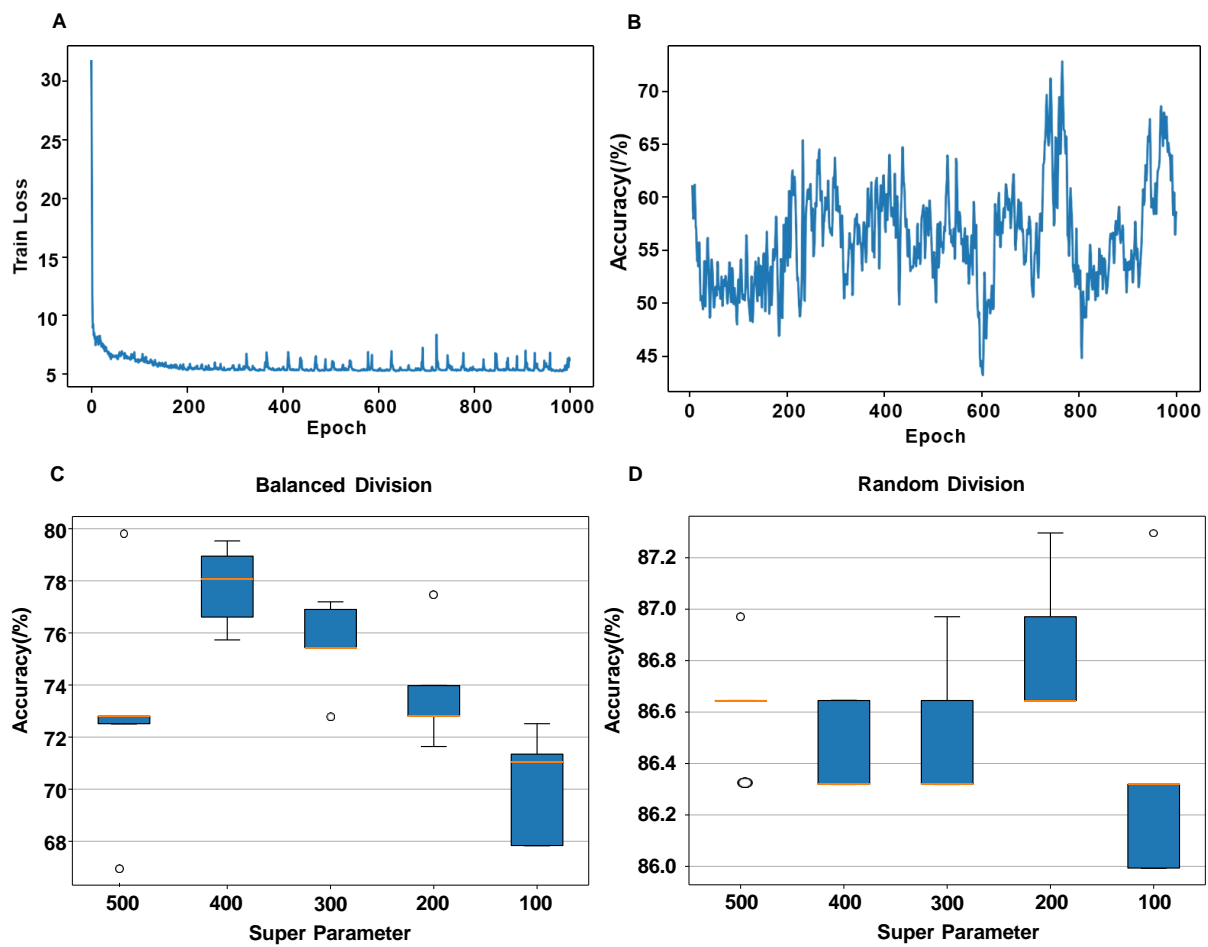

**Figure 2. Results of HDDN**
(A) Training loss of HDDN with epoch increasing.
(B) Accuracy in test-set of HDDN with epoch increasing.
(C and D) Average accuracy in different super parameters is close to each other, which indicates our model is insensitive to super parameters.

As we present above, test-set in Balanced Dataset contains 50%incompatible-50%compatible samples, so the random or ineffective model can reach up to at least 50% accuracy. After trained and tested five times, our model reaches 75.75% average accuracy on test-set, which is obviously higher than other models (HSP:68.93%; MLP:63.56%; CDN:65.97%; DN:62.42%). What's more, our model obtains higher precision (76.99%), recall (74.13%), specificity (77.31%) and $F_1$ score (74.80%) than all other models. The results show that our architecture design performs not only better on random division dataset but also better on strict division dataset, and really discovers the mystery of polymer blend compatibility. At the same time, our model is super-parameter insensitive, which is shown in Figure 2.

We also find that all models get lower scores than on Random Dataset, and the strict division may be the main reason. In the Random Dataset, training-set contains the same kinds of polymer blend combination in test-set. Although composition rate may be different, the influence of composition is obviously lower than that of structure representation. Therefore, model can easily handle such data in test-set according the experience of training-set. However, In Balanced Dataset, training-set and test-set are strictly separated so that imitating is hard because of the lack of direct experience. This phenomenon also explains why it get lower accuracy when we train it on a strict division dataset.

**Ablation Results**
We are interested in whether each module we design works, so we conduct ablation experiments to show the necessity of each module. Ablation experiment is presented by Ren [51], and the principle is similar to variant-control in biology and chemistry. We remove some modules from our model or replace them with other connections. In this process, we will prove that our modules are working together well and all make contributions to this predictive task. We present the detail of HDDN as Figure 3, and name these incomplete models HDDN-noc, HDDN-nodense, HDDN-nodiff and HDDN-noabs.

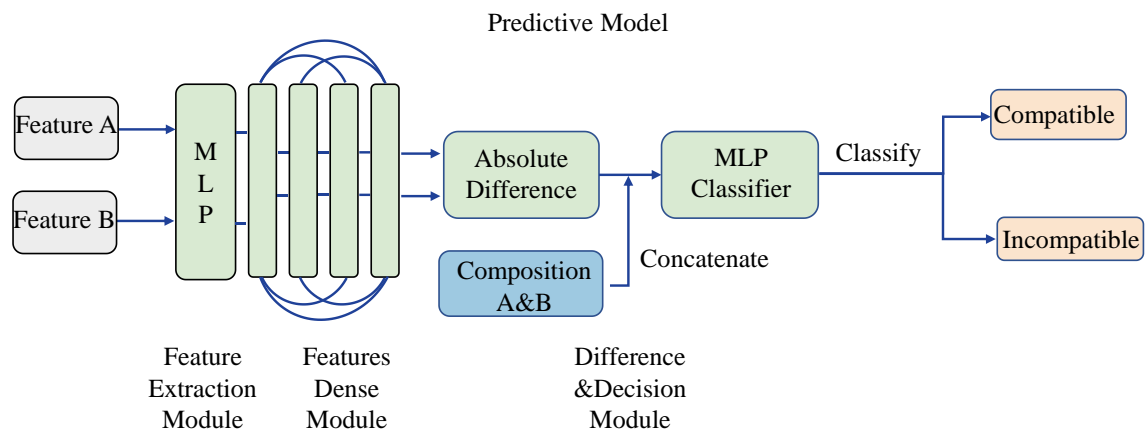

**Figure 3. Structure of HDDN**
Feature Extraction: reduce the dimension of sparse input and extract features. Feature Dense Connection: connect all hidden layers, so that the output of the last layer can integrate different levels of molecular structure information. Difference and Decision: make the difference between the two polymer structure vectors and use MLP classifier to get compatible decision.

- HDDN-noc: we dropout composition to observe the performances for the cases without composition information.
- HDDN-nodense: we replace the dense connection by MLP and keep the number of layers and nodes same, in order to investigate the effect of dense connection.
- HDDN-nodiff: we remove the difference module and concatenate two polymer vector representation as input, so the 4096-D vector get through dense connection and come into decision module directly.
- HDDN-noabs: we remove the absolute value process and use difference directly, in order to find out whether absolute process is useful.

**Table 3. Results on Test-set in Ablation Experiments**

| Model | Random Division Results | | | | | |
|---|---|---|---|---|---|---|
| | MSE | Accuracy (%) | Precision (%) | Recall (%) | Specificity (%) | $F_1$ score (%) |
| HDDN-noc | 8.26 | 88.89 | 87.34 | 84.68 | 91.69 | 85.86 |
| HDDN-nodense | 6.68 | 90.43 | 90.07 | 85.59 | 93.64 | 87.71 |
| HDDN-nodiff | **4.48** | 89.53 | **91.41** | 81.42 | **94.91** | 86.12 |
| HDDN-noabs | 5.42 | 88.99 | 89.87 | 81.64 | 93.86 | 85.54 |
| HDDN | 7.25 | **90.89** | 89.75 | **87.24** | 93.31 | **88.43** |
| Model | Balanced Division Results | | | | | |
| | MSE | Accuracy (%) | Precision (%) | Recall (%) | Specificity (%) | $F_1$ score (%) |
| HDDN-noc | 6.26 | 70.51 | 67.30 | 78.33 | 62.98 | 72.09 |
| HDDN-nodense | 6.37 | 70.10 | 67.30 | **78.33** | 62.98 | 71.07 |
| HDDN-nodiff | **5.81** | 62.30 | 66.72 | 46.25 | **77.72** | 54.53 |
| HDDN-noabs | 4.60 | 60.09 | 57.26 | 75.83 | 44.95 | 65.03 |
| HDDN | 6.14 | **75.75** | **76.99** | 74.13 | 77.31 | **74.80** |

Results of ablation experiments are listed in Table 3. On random division dataset, we can find that HDDN gets the highest accuracy (90.89%) and at the same time, all alternatives perform well (gaps are smaller than 2%). As for the attention on compatible and incompatible samples, alternatives all get higher precision and specificity than HDDN and lower recall and $F_1$ score than HDDN. This is because that defect in architecture may cause model pays more attention on samples themselves but not the mechanism. HDDN get the highest $F_1$ score and this indicates that HDDN can balance precision and recall in the best way.

On Balanced division dataset, complete HDDN model reaches highest accuracy, and other alternatives get obviously lower accuracy than HDDN (75.75%). As for $F_1$ score, HDDN is also obviously higher than other alternatives. In details, we can see two thread polymer representation and difference module is the most important part of the whole architecture (HDDN-nodiff and HDDN-noabs get low accuracy). This is because with two threads operation, we guarantee the position invariance between two polymers and we needn't worry about the turn where we put two representations together. At the same time, our difference method is related with the chemistry theory, and proved to be effective. As for HDDN-noc, we can learn that composition rate is also very important to our model, which is also consistent to the truth that we can not only predict compatibility without composition. Without dense connection or absolute value module, HDDN will lose some representation ability and cannot learn the true mechanism well through training. All analyses above are also true when we turn to $F_1$ score. From results and analyses of ablation experiments above, we can infer that each module in our model is useful and necessary. They work together effectively on solving this classification problem.

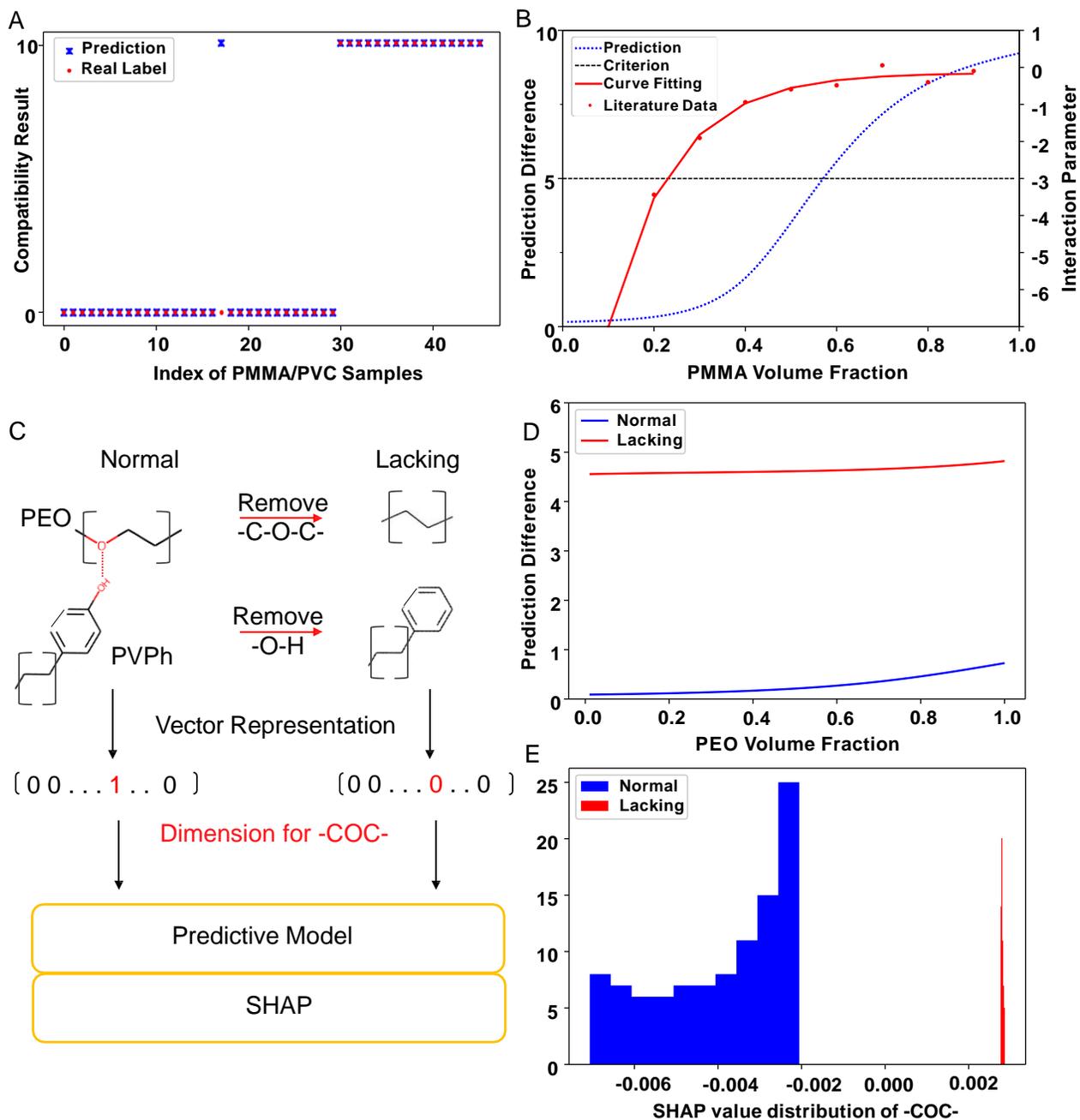

**Figure 4 Case Study of PMMA/PVC and Interpretability HDDN with PEO/PVPh**
(A) Our prediction to PMMA/PVC blend samples and real label of it. Different samples have different composition and share the same structure vector.
(B) The blue dotted line indicates the relationship between composition and our prediction difference for PMMA/PVC polymer blend. The red solid line indicates the relationship between composition and the Flory–Huggins interaction parameter. The horizontal line is our predictor criterion.
(C) Structures of Normal PEO/PVPh polymer blend and Lacking one are shown. Dimension for -COC- is specific and highlighted. Normal and Lacking both pass through the predictive model and SHAP model.
(D) The relationship between composition and our prediction difference for Normal and Lacking PEO/PVPh polymer blend.
(E) SHAP value distribution of -COC- dimension of Normal PEO/PVPh and Lacking PEO/PVPh.

### Confidence Tests

As we mentioned above, we will make a judgement that the model works if it can get higher accuracy than random classification. We will now discuss at least how accurate we are confident the model will be. We can assume that there is an objective accuracy rate behind each model architecture method. We will mainly discuss and prove our confidence in the accuracy of HDDN model.

Based on our hypothesis, the prediction behavior is a Bernoulli trail, and we define the probability of success as $\theta$. In Balanced Division, we test our model on 306 samples for five times, which corresponds to Binomial trail. And we define that the correct classification numbers $X$ follows $B(n, \theta)$, where $n = 306*5 = 1530$:

$$X \sim B(n, \theta), \qquad P(X = x) = \binom{n}{x} \theta^x (1-\theta)^{(n-x)} \tag{3}$$

where observation of $X$ is $X_0$. We make hypothesis as follow:

$$H_0: \theta = \theta_0 \quad V.S. \quad H_1: \theta > \theta_0. \tag{4}$$

For hypothesis test, we conduct P-value confidence test:

$$P\ value = \sup_{\theta=\theta_0} P_\theta(X > X_0) = \sup_{\theta=\theta_0} \sum_{i=X_0}^{n} \binom{n}{i} \theta^i (1-\theta)^{(n-i)}. \tag{5}$$

In our experiments, we reach 75.75% average accuracy and we can substitute $X_0 = 1530 * 75.75\% = 1159$ into the formula. After calculation, when $\theta_0$ is equal to 73.07% and 73.87%, the P-values are equal to 0.01 and 0.05 respectively. That's to say, we can state that our model accuracy is higher than 73.07% at a significance level no less than 0.01 and is higher than 73.87% at a significance level no less than 0.05 while the significance levels of 0.01 or 0.05 is conservative. Based on the above reasoning, we prove that our model really learns how the structure-property mechanism works after training.

**Case Study and Interpretability**

Following, we will pay attention to specific case to verify how our method performs on polymer blend compatibility.

The poly(methyl methacrylate) (PMMA) / poly(vinyl chloride) (PVC) blend is widely used as a polymer electrolyte. The poor mechanical flexibility of the PMMA film limits its use in energy storage devices, while the mix of PVC can improve the mechanical and electrical properties of the polymer electrolyte, and the conductivity as well [52,53]. In contrast, PMMA is also considered as a processing aid in PVC production process. PMMA helps in the plasticization of PVC and helps both in the processing of PVC and constituting a blend material.

We address the compatibility prediction of PMMA/PVC with our ML method. We use the model trained by Balanced Division, and utilize all 46 entries of PMMA/PVC in our database. These entries share the same polymer representation but different composition. We find that our model only misses one single sample and achieve accurate results in all other ones, which proves the effectiveness of our method. As mentioned before, the Flory-Huggins parameter of the polymers and the intermolecular interactions between different repeat units are important factors that determine the mixed condition. Existing research shows that with PMMA volume fraction increasing, $\chi_{12}$ rises from negative values to zero, which indicates that compatibility is falling [54]. Through our ML method, prediction difference also rises with PMMA volume fraction increasing, and gradually more than the criterion value. It shows the consistency between our model and experimental results in the tendency. As for the bias between two curves in the range of 0% to 60%, it can be due to the inequacy of our data and the imperfection of our model. In the whole, we can find that our model is truly efficient to polymer compatibility prediction.

Our method has been proved to be effective to polymer compatibility prediction, while we want to go further and investigate whether this model can be interpreted with chemistry knowledge. Based on our design, every dimension represents specific chemistry structure and the relation can be obtained with RDkit Python Package. During several approaches to interpreting model prediction, we utilize SHAP (SHapley Additive exPlanations) [55] to estimate influence weight of each dimension.

We choose PEO/PVPh (poly(ethylene oxide)/poly(p-hydroxystyrene)) blend and apply the interpreting method to it. The process and result are shown in Figure 4C. According to chemistry analysis, we find that the compatibility of PEO/PVPh system is increased due to the hydrogen bonding forces of the hydroxyl groups [56]. To verify the consistency between our model and chemistry knowledge, we set special Lacking PEO/PVPh by removing -COC- of PEO and -OH of PVPh. In addition, we choose the special dimension which refers to the existence of ether bond (-COC-) and calculate its value. Through our model, we find that prediction difference of Lacking PEO/PVPh is much higher than Normal one, which means that Lacking is obviously more incompatible than Normal. As for SHAP values of -COC-, we find that values from Normal is distributed more negative. It can be referred that, after regular sampling, existence of -COC- tends to take negative influence to prediction and make model assume blend more compatible. These phenomena prove that our model can be interpreted with chemistry knowledge.

**DISCUSSION**

In this work, we present a general ML method for material property prediction, and choose compatibility as our focus. To address this task, we establish a dataset based on literature mining and NIMS. We show that our model indeed achieves impressive classification results and performs better than chemistry methods and other possible competing ML models. Through ablation experiments we explain why our architecture design can work and quantify the contribution of each module. Each module in HDDN works for specific purpose and effectively contributes to the prediction task. Furthermore, we conduct confidence test and case study to demonstrate the reliability of our model. Interpretability investigation proves that our model can be interpreted with chemistry knowledge, and more details of our model can be investigated as supplement to existing chemistry knowledge.

We also get some other interesting findings in our research. As to compatibility, our method pays more attention to polymer structure and cares less about composition. HDDN model successfully deals with the compatibility prediction of binary blends, but it cannot be directly used to deal with the compatibility of ternary or more blends and copolymers (such as ABS (Acrylonitrile Butadiene Styrene)), which needs to be further expanded. In the future, we will try to optimize our model structure though it may be challenging. We will also apply this general method to other significant scientific problems, such as polymer self-assembly, biodegradable materials and so on.

## EXPERIMENTAL PROCEDURES
### Resource Availability

*Lead Contact*
Further information and requests for resources and reagents should be directed to and will be fulfilled by the Changshui Zhang (zcs@mail.tsinghua.edu.cn).

*Materials Availability*
This study did not generate new, unique reagents.

*Data and Code Availability*
The data and main code to reproduce the results of this study are available at following GitHub page:
https://github.com/Zhilong-Liang/Polymer-Compatibility

### Machine-learning Experiment Settings
All experiments are implemented in Python using Pytorch toolbox [57], and the computing is accelerated on a NVIDIA GeForce RTX 2080 Ti GPU. We set the maximal training epoch number to 1000 and the mini-batch to 20. The initial learning rate is set to $10^{-4}$ for Balanced Division and $5*10^{-5}$ for Random Division. Loss related parameter $\lambda$ is set as 10, which determines the label of incompatible samples. Total time spend is about 10 min for a complete training of 1000 iterations. Adam optimizer is used to optimize the loss function. As for criterion, we choose MSE (Mean Square Error) as the loss function for the phase.

Since we are accomplishing a classification prediction task, we also care about the accuracy of the model. In addition, we test some other indices:

$$Accuracy = (TP + TF)/(P + F), \quad (6a)$$
$$Precision = TP/(TP + FP), \quad (6b)$$
$$Recall = TP/(TP + FN), \quad (6c)$$
$$Specificity = TN/(TN + FP), \quad (6d)$$
$$F_1\ score = 2 \times \frac{Precision \times Recall}{Precision + Recall}, \quad (6e)$$

where $P$ means incompatible, $N$ means compatible, $T$ means true, and $F$ means false. Precision refers to the proportion of the "incompatible" predictions that are actually correct. Recall refers to the proportion of incompatible samples with correct prediction among all incompatible samples, which is numerically equal to Sensitivity and both indicate the ability to predict incompatible cases. Specificity means the ability to predict compatible cases. The $F_1$ score comprehensively considers the effects of precision and recall. If one of them is too small, the value of $F_1$ will become smaller.

## SUPPLEMENTAL INFORMATION
Document S1 Supplemental Machine Learning Method.

## ACKNOWLEDGMENTS
The authors are grateful for support from National Institute for Materials Science. We would like to thank Shuojin Wang et al. for the assistance in processing data. Meanwhile, we also would like to thank Haodi Liu for polishing the language and Weisheng Pan for advice in interpretability study.

## AUTHOR CONTRIBUTIONS
Z. Liang, C. Zhang. and J. Yuan conceived of the main research idea. Z. Liang carried out method design, machine-learning modeling and wrote the manuscript. Z. Li and S. Zhou took part in the literature summary and conducted the HSP method experiments. Y. Sun took part in the experiment design. C. Zhang and J. Yuan supervised the project and revised the manuscript. All authors discussed the results and commented on the manuscript.

## DECLARATION OF INTERESTS
The authors declare no competing interests.